\documentclass{article}
\usepackage{ijcai21}

\usepackage{times}
\usepackage{soul}
\usepackage{url}
\usepackage[hidelinks]{hyperref}
\usepackage[utf8]{inputenc}
\usepackage[small]{caption}
\usepackage{graphicx}
\usepackage{amsmath}
\usepackage{amsthm}
\usepackage{booktabs}
\usepackage{algorithm}
\usepackage{algorithmic}
\usepackage{amssymb}
\usepackage{xcolor}

\usepackage{amsfonts}
\usepackage{bbm}
\usepackage{multirow}
\usepackage{graphicx}
\usepackage{float}
\usepackage{subfigure} 

\title{Unsupervised Hashing with Contrastive Information Bottleneck}
\author{
Zexuan Qiu$^1$\and
Qinliang Su$^1$\footnote{Corresponding author. Qinliang Su is also affiliated with (\textit{i}) Guangdong Key Laboratory of Big Data Analysis and Processing, Guangzhou, China, and (\textit{ii}) Key Laboratory of Machine Intelligence and Advanced Computing, Ministry of Education, China. }\and
Zijing Ou$^{1}$\and
Jianxing Yu$^2$\And 
Changyou Chen$^3$\\
\affiliations
$^1$School of Computer Science and Engineering, Sun Yat-sen University, Guangzhou, China\\
$^2$School of Artificial Intelligence, Sun Yat-sen University, Guangdong, China\\
$^3$CSE Department, SUNY at Buffalo\\
\emails
\{qiuzx2, ouzj\}@mail2.sysu.edu.cn, 
\{suqliang, yujx26\}@mail.sysu.edu.cn, 
changyou@buffalo.edu 
}

\begin{document}

\maketitle

\begin{abstract}


Many unsupervised hashing methods are implicitly established on the idea of reconstructing the input data, which basically encourages the hashing codes to retain as much information of original data as possible. However, this requirement may force the models spending lots of their effort on reconstructing the unuseful background information, while ignoring to preserve the discriminative semantic information that is more important for the hashing task. To tackle this problem, inspired by the recent success of contrastive learning in learning continuous representations, we propose to adapt this framework to learn binary hashing codes. Specifically, we first propose to modify the objective function to meet the specific requirement of hashing and then introduce a probabilistic binary representation layer into the model to facilitate end-to-end training of the entire model. We further prove the strong connection between the proposed contrastive-learning-based hashing method and the mutual information, and show that the proposed model can be considered under the broader framework of the information bottleneck (IB). Under this perspective, a more general hashing model is naturally obtained. Extensive experimental results on three benchmark image datasets demonstrate that the proposed hashing method significantly outperforms existing baselines\footnote{Our code is available at \url{https://github.com/qiuzx2/CIBHash}.}.

\end{abstract}

\section{Introduction}

In the era of big data, similarity search, also known as approximate nearest neighbor search, plays a pivotal role in modern information retrieval systems, like content-based image and document search, multimedia retrieval and plagiarism detection \cite{lew2006content}, etc. If the search is carried out directly in the original real-valued feature space, the cost would be extremely high in both storage and computation due to the huge amount of data items. On the contrary, by representing every data item with a compact binary code that preserves the similarity information of items, the hashing technique can significantly reduce the memory footprint and increase the search efficiency by working in the binary Hamming space, and thus has attracted significant attention in recent years.


Existing hashing methods can be roughly categorized into supervised and unsupervised groups. Although supervised hashing generally performs better due to the availability of label information during training, unsupervised hashing is more favourable in practice. Currently, many of competitive unsupervised hashing methods are established based on the goal of satisfying data reconstruction. For instances, in \cite{do2016learning}, binary codes are found by solving a discrete optimization problem on the error between the reconstructed and original images. Later, to reduce the computational complexity, variational auto-encoders (VAE) are trained to directly output hashing codes \cite{DaiGKHS17,ShenLS19}, where the decoder is enforced to reconstruct the original images from the binary codes. Recently, the encoder-decoder structure is used in conjunction with the graph neural network to better exploit the similarity information in the original data \cite{ShenQCY00S020}. A similar idea is also explored under the generative adversarial network in \cite{SongHGXHS18}. It can be seen that what these methods really do is to force the codes to retain as much information of original data as possible through the reconstruction constraint. However, in the task of hashing, the objective is not to retain all information of the original data, but to preserve the distinctive similarity information. For example, the background of an image is known to be unuseful for similarity search. But if the reconstruction criterion is employed, the model may spend lots of effort on reconstructing the background, while ignoring the preservation of more meaningful similarity information.

Recently, a non-reconstruction-based unsupervised representation learning framework ({\textit{i.e.}}, contrastive learning) is proposed \cite{ChenK0H20}, which learns real-valued representations by maximizing the agreement between different views of the same image. It has been shown that the method is able to produce semantic representations that are comparable to those obtained under supervised paradigms. Inspired by the tremendous success of contrastive learning, in this paper, we propose to learn the binary codes under the non-reconstruction-based contrastive learning framework. However, due to the continuous characteristic of learned representations and the mismatch of objectives, the standard contrastive learning does not work at its best on the task of hashing. Therefore, we propose to adapt the framework by modifying its original structure and introducing a probabilistic binary representation layer into the model. In this way, we can not only bridge objective mismatching, but also train the binary discrete model in an end-to-end manner, with the help from recent advances on gradient estimators for functions involving discrete random variables \cite{BengioLC13,MaddisonMT17}. Furthermore, we establish a connection between the proposed probabilistic hashing method and mutual information, and show that the proposed contrastive-learning-based hashing method can be considered under the broader information bottleneck (IB) principle \cite{tishby2000information}. Under this perspective, a more general probabilistic hashing model is obtained, which not only minimizes the contrastive loss but also seeks to reduce the mutual information between the codes and original input data. Extensive experiments are conducted on three benchmark image datasets to evaluate the performance of the proposed hashing methods. The experimental results demonstrate that the contrastive learning and information bottleneck can both contribute significantly to the improvement of retrieval performance, outperforming the state of the art (SOTA) by significant margins on all three considered datasets.

\section{Related Work}

\paragraph{Unsupervised hashing} 
Many of unsupervised hashing methods are established on the generative architectures. Roughly, they can be divided into two categories. On the one hand, some of them adopt the encoder-decoder architecture \cite{DaiGKHS17,zheng2020generative,dong2019document,ShenQCY00S020,ShenLS19} and seek to reconstruct the original images or texts. For example, SGH \cite{DaiGKHS17} proposes a novel generative approach to learn stochastic binary hashing codes through the minimum description length principle. TBH \cite{ShenQCY00S020} puts forward a variant of Wasserstein Auto-encoder with the code-driven adjacency graph to guide the image reconstruction process. On the other hand, some models employ generative adversarial nets to implicitly maximize reconstruction likelihood through the discriminator \cite{SongHGXHS18,DizajiZSYDH18,ZiebaSET18}. However, by reconstructing images, these methods could introduce overload information into the hashing code and therefore hurt the model's performance. In addition to the reconstruction-based methods, there are also some hashing models that construct the semantic structure based on the similarity graph constructed in the original high-dimensional feature space. For instance, DistillHash \cite{YangLDLT19} addresses the absence of supervisory signals by distilling data pairs with confident semantic similarity relationships. SSDH \cite{YangDLLT18} constructs the semantic structure through two half Gaussian distributions. Among these hashing methods, DeepBit \cite{LinLCZ16} and UTH \cite{HuangXZW17} are somewhat similar to our proposed model, both of which partially consider different views of images to optimize the hashing codes. However, both methods lack a carefully and theoretically designed framework, leading to poor performance.

\paragraph{Contrastive Learning}

Recently, contrastive learning has gained great success in unsupervised representation learning domains. SimCLR \cite{ChenK0H20} proposes a simple self-supervised learning network without requiring specialized architectures or a memory bank, but still achieves excellent performance on ImageNet. MoCo \cite{He0WXG20} builds a dynamic and consistent dictionary preserving the candidate keys to enlarge the size of negative samples. BYOL \cite{GrillSATRBDPGAP20} abandons negative pairs by introducing the online and target networks.

\paragraph{Information Bottleneck}
The original information bottleneck (IB) work \cite{tishby2000information} provides a novel principle for representation learning, claiming that a good representation should retain useful information to predict the labels while eliminating superfluous information about the original sample. Recently, in \cite{AlemiFD017}, a variational approximation to the IB method is proposed. \cite{Federici0FKA20} extends the IB method to the representation learning under the multi-view unsupervised setting.

\section{Hashing via Contrastive Learning}

In this section, we will first give a brief introduction to the contrastive learning, and then present how to adapt it to the task of hashing.

\subsection{Contrastive Learning}
Given a minibatch of images $x^{(k)}$ for $i=1,2, \cdots, N$, the contrastive learning first transforms each image into two views $v_1^{(k)}$ and $v_2^{(k)}$ by applying a combination of different transformations (\textit{e.g.}, cropping, color distortion, rotation, etc) to the image $x^{(k)}$. After that, the views are fed into an encoder network $f_\theta(\cdot)$ to produce continuous representations
\begin{equation}
z_i^{(k)} = f_\theta(v_i^{(k)}),   \quad i=1 \;\text{or} \;2.
\end{equation}
Since $z_1^{(k)}$ and $z_2^{(k)}$ are derived from different views of the same image $x^{(k)}$, they should mostly contain the same semantic information. The contrastive learning framework achieves this goal by first projecting $z_i^{(k)}$ into a new latent space with a projection head
\begin{equation}
h_i^{(k)} = g_\phi(z_i^{(k)})
\end{equation}
and then minimizing the contrastive loss\footnote{In this paper, we adopt NT-Xent Loss \cite{ChenK0H20} as the contrastive loss function.} on the projected vectors $h_i^{(k)}$ as
\begin{equation}
L_{cl} = \frac{1}{N} \sum_{k=1}^N \left(\ell_1^{(k)} + \ell_2^{(k)}\right),
\end{equation}
where 
\begin{equation}
\ell_1^{(k)} \! \triangleq -\log \frac{e^{{sim(h_1^{(k)},h_2^{(k)})}/{\tau}}} 
{e^{{sim(h_1^{(k)},h_2^{(k)})}/{\tau}}   \! + \! \sum\limits_{i, n \neq k}e^{{sim(h_1^{(k)},h_i^{(n)})}/{\tau}}},
\label{fml:singleloss}
\end{equation}
and $sim(h_1, h_2) \triangleq  \frac{h_1^Th_2}{{\left\|{h_1}\right\| \left\|h_2\right\|}}$ means the cosine similarity; and $\tau$ denotes a temperature parameter controlling the concentration level of the distribution \cite{HintonVD15}; and $\ell^{(k)}_2$ can be defined similarly by concentrating on $h_2^{(k)}$. In practice, the projection head $g_\phi(\cdot)$ is constituted by an one-layer neural network. After training, for a given image $x^{(k)}$, we can obtain its representation by feeding it into the encoder network
\begin{equation}
r^{(k)} = f_\theta(x^{(k)}).
\end{equation}
Note that $r^{(k)}$ is different from $z^{(k)}_1$ and $z^{(k)}_2$ since $r^{(k)}$ is extracted from the original image $x^{(k)}$ rather than its views. The representations are then used for downstream applications.

\subsection{Adapting Contrastive Learning to Hashing}
\label{AdaptingCL2Hash}

To obtain the binary hashing codes, the simplest way is to set a threshold on every dimension of the latent representation to binarize the continuous representation like
\begin{equation}
[b^{(k)}]_d =
\begin{cases}
0,&  [r^{(k)}]_d < [c]_d\\
1,&  [r^{(k)}]_d \geq [c]_d
\end{cases}
\end{equation}
where $[\cdot]_d$ denotes the $d$-th element of a vector; and $c$ is the threshold vector. There are many different ways to decide the threshold values. For example, we can set $[c]_d$ to be $0$ or the median value of all representations on the $d$-th dimension to balance the number of $0$'s and $1$'s \cite{BalujaC08}. Although it is simple, two issues hinder this method from releasing its greatest potential. {1) \emph{Objective Mismatch}}: The primary goal of contrastive learning is to extract representations for downstream {\emph{discriminative}} tasks, like classification or classification with very few labels. However, the goal of hashing is to preserve the semantic similarity information in the binary representations so that similar items can be retrieved quickly simply according to their Hamming distances. {2) \emph{Separate Training}}: To obtain the binary codes from the representations of original contrastive learning, an extra step is required to binarize the real-valued representations. Due to the non-differentiability of binarization, the operation can not be incorporated for joint training of the entire model.

For the objective mismatch issue, by noticing that the contrastive loss is actually defined on the similarity metric, we can simply drop the projection head and apply the contrastive loss on the binary representations directly. For the issue of joint training, we follow the approach in \cite{shen2018nash} by introducing a probabilistic binary representation layer into the model. Specifically, given a view $v_i^{(k)}$ from the $k$-th image $x^{(k)}$, we first compute the probability
\begin{equation}
p_i^{(k)}  = \sigma(z_i^{(k)}),
\end{equation}
where $z_i^{(k)} = f_\theta(v_i^{(k)})$ is the continuous representation; and $\sigma$ denotes the sigmoid function and is applied to its argument in an element-wise way. Then, the binary codes are generated by sampling from the multivariate Bernoulli distribution as
 \begin{equation}
b^{(k)}_i \sim Bernoulli(p_i^{(k)}),
\end{equation}
where the $d$-th element $[b^{(k)}_i]_d$ is generated according to the corresponding probability $[p^{(k)}_i]_d$. Since the binary representations $b_i^{(k)}$ are probabilistic, to have them preserve as much similarity information as possible, we can minimize the expected contrastive loss
\begin{equation}
\bar L_{cl} = \frac{1}{N} \sum_{k=1}^N\left(\bar \ell_1^{(k)} + \bar \ell_2^{(k)} \right),
\end{equation}
where
\begin{equation} \label{contrastive_loss_binary}
\bar \ell_{1}^{(k)} \!=\! - {\mathbb{E}} \!\!\left[\!\log \! \frac{e^{{sim(b_1^{(k)},b_2^{(k)})}/{\tau}}} 
{e^{{sim(b_1^{(k)}, b_2^{(k)})}/{\tau}}   \!\! + \!\!\!\! \sum\limits_{i, n \neq k} \!\! e^{{sim(b_1^{(k)}, b_i^{(n)})}/{\tau}}}\right];
\end{equation}
$b_i^{(k)} \sim Bernoulli\left(\sigma(f_\theta(v_i^{(k)}))\right)$ and the expectation ${\mathbb{E}} [\cdot]$ is taken w.r.t. all $b_i^{(k)}$ for $i=1,2$ and $k=1,2, \cdots, N$; and $\bar \ell_{2}^{(k)}$ can be defined similarly. Note that the contrastive loss is applied on the binary codes $b^{(k)}_i$ directly. This is because the similarity-based contrastive loss without projection head is more consistent with the objective of the hashing task. 

The problem now reduces to how to minimize the loss function $\bar L_{cl}$ {\textit{w.r.t.}} the model parameters $\theta$, the key of which lies in how to efficiently compute the gradients $\frac{\partial \bar \ell_1^{(k)}}{\partial \theta}$ and $\frac{\partial \bar \ell_2^{(k)}}{\partial \theta}$. Recently, many efficient methods have been proposed to estimate the gradient of neural networks involving discrete stochastic variables \cite{BengioLC13,MaddisonMT17}. In this paper, we employ the simplest one, the straight-through (ST) gradient estimator \cite{BengioLC13}, and leave the use of more advanced estimators for future exploration. Specifically, the ST estimator first reparameterizes $b_i^{(k)}$ as
\begin{align}
\widetilde b_i^{(k)} = \frac{\mathrm{sign}(\sigma(f_\theta(v_i^{(k)})) - u) + 1}{2},
\end{align}
and use the reparameterized $\widetilde b_i^{(k)}$ to replace $b_i^{(k)}$ in \eqref{contrastive_loss_binary} to obtain an approximate loss expression, where $u$ denotes a sample from the uniform distribution $[0, 1]$. The gradient, as proposed in the ST estimator, can then be estimated by applying the backpropagation algorithm on the approximate loss. In this way, the entire model can be trained efficiently in an end-to-end manner. When the model is used to produce binary code for an image $x$ at the testing stage, since we want every image to be deterministically associated with a hashing code, we drop the effect of randomness in the training stage and produce the code by simply testing whether the probability $\sigma\left(f_\theta(x)\right)$ is larger than 0.5 or not.

\section{Improving under the IB Framework}
In this section, we first establish the connection between the probabilistic hashing model proposed above and the mutual information, and then reformulate the learning to hashing problem under the broader IB framework.

\subsection{Connections to Mutual Information}
For the convenience of presentation, we re-organize the minibatch of views $\{v_1^{(k)}, v_2^{(k)}\}_{k=1}^N$ into the form of $\{v_i\}_{i=1}^{2N}$. We randomly take one view  ({\textit{e.g.}}, $v_k$) as the target view for consideration. For the rest of $2N-1$ views $\{v_i\}_{i\ne k}$, we assign each of them a unique label from $\{1, 2, \cdots, 2N-1\}$ according to some rules. The target view $v_k$ is assigned the label same as the view derived from the same image. Without loss of generality, we denote the label as $y_k$. Now, we want to train a classifier to predict the label of the target view $v_k$. But instead of using the prevalent softmax-based classifiers, we require the classifier to be memory-based. Specifically, we extract a stochastic feature  $b_i \sim Bernoulli\left(\sigma(f_\theta(v_i))\right)$ for each view in the minibatch and predict the probability that $v_k$ belongs to label $y_k$ as
\begin{equation}
   p(y_k|v_k, {\mathcal{V}}) = \frac{e^{{sim(b_k, {\mathcal{B}}{\backslash b_k} (y_k))}/{\tau}}} 
   {  \sum_{{{c}}\in {\mathcal{B}} \backslash b_k} e^{{sim(b_k, c)}/{\tau}}},
\end{equation}
where ${\mathcal{V}} \triangleq \{v_1, v_2, \cdots, v_{2N}\}$ is the set of views in the considered minibatch; and ${\mathcal{B}} \triangleq \{b_1, b_2, \cdots, b_{2N}\}$ is the set of stochastic features derived in the minibatch, while ${\mathcal{B}} \backslash b_k$ means the set that excludes the element $b_k$ from ${\mathcal{B}}$; and ${\mathcal{B}}{\backslash b_k} (y)$ denotes the feature that is associated with the label $y$ in the set ${\mathcal{B}} \backslash b_k$ .
Since the features $b_i\in {\mathcal{B}}$ are stochastic, we take expectation over the log-probability above and obtain the loss {\textit{w.r.t.}} the view-label pair $(v_k, y_k)$ under the considered minibatch ${\mathcal{V}}$ as
\begin{equation} \label{cross-entropy}
\ell_{ce}(v_k, y_k) \!=\! -{\mathbb{E}}_{p({\mathcal{B}}|{\mathcal{V}})} \! \left[\log \frac{e^{{sim(b_k, {\mathcal{B}}{\backslash b_k} (y_k))}/{\tau}}} 
{  \sum_{{{c}}\in {\mathcal{B}} \backslash b_k} e^{{sim(b_k, c)}/{\tau}}} \right],
\end{equation}
where $p({\mathcal{B}}| {\mathcal{V}}) = \prod_{i=1}^{2N} {\mathbb{P}}(b_i|v_i)$ with 
\begin{equation}
{\mathbb{P}} (b|v) \triangleq Bernoulli\left(\sigma(f_\theta(v))\right).
\end{equation}
By comparing \eqref{cross-entropy} to \eqref{contrastive_loss_binary}, we can see that the two losses are actually the same. Therefore, training the proposed probabilistic hashing model is equivalent to minimizing the cross-entropy loss of the proposed memory-based classifier.

The loss function in \eqref{cross-entropy} is only responsible for the $k$-th view under one minibatch. For the training, the loss should be optimized over a lot of view-label pairs from different minibatches. Without loss of generality, we denote the distribution followed by view-label pairs as ${{\mathbb{P}}(v, y)}$. Then, we can express the loss averaged over all view-label pairs as
\begin{equation}
L_{ce} = - {\mathbb{E}}_{{\mathbb{P}}(v, y)} {\mathbb{E}}_{p({\mathcal{B}}| {\mathcal{V}} )}\left[\log q(y|b)\right],
\end{equation}
where the distribution $ q(y|b)$ is defined as
\begin{equation}
  q(y|b) = \frac{e^{{sim(b, {\mathcal{B}}{\backslash b} (y))}/{\tau}}} 
  {  \sum_{{{c}}\in {\mathcal{B}} \backslash b} e^{{sim(b, c)}/{\tau}}};
\end{equation}
and ${\mathcal{V}}$ represents the minibatch of views that $(v, y)$ is associated with. Without affecting the final conclusions, for the clarity of analysis, we only consider the randomness in $b \sim {\mathbb{P}}(b|v)$, while ignoring the randomness in ${\mathcal{B}}\backslash b$. Under this convention, the loss $L_{ce}$ can be written as
\begin{align} \label{Lce_lower_bound}
L_{ce} &= -\int {{\mathbb{P}}(b, y) \log q(y|b) dy db},
\end{align}
where ${\mathbb{P}}(b, y) = \int{{\mathbb{P}}(v, y){\mathbb{P}}(b|v)dv}$. From the inequality $\int {{\mathbb{P}}(b, y) \log {\mathbb{P}}(y|b) dy db} \ge \int {{\mathbb{P}}(b, y) \log q(y|b) dy db} $, which can be easily derived from the non-negativeness of KL-divergence, it can be seen that
\begin{align}
 L_{ce} \ge  -\int {{\mathbb{P}}(b, y) \log {\mathbb{P}}(y|b) dy db}  = H(Y|B),
\end{align}
where ${\mathbb{P}}(y|b)$ denotes the conditional distribution from the joint pdf ${\mathbb{P}}(b, y)$; $B$ and $Y$ denotes the random variables of binary representations and labels, whose marginal distributions are ${\mathbb{P}}(b)$ and ${\mathbb{P}}(y)$, respectively. Then, subtracting entropy $H(Y)$ on both sides gives
\begin{equation}
L_{ce} - H(Y) \ge - I(Y, B),
\end{equation}
in which the equality $I(Y, B)=H(Y) - H(Y|B)$ is used. Therefore, $L_{ce} - H(Y)$ is actually an upper bound of the negative mutual information $- I(Y, B)$. Because the entropy of labels $H(Y)$ is a constant, minimizing the loss function $L_{ce}$ is equivalent to minimizing the upper bound of negative mutual information. Thus, we can conclude that the proposed model essentially maximizes the mutual information, {\textit{i.e.}}, 
\begin{equation}
\max_\theta  I(Y, B),
\end{equation}
between binary representations $B$ and labels $Y$ under the joint probabilistic model ${\mathbb{P}}(v, y, b) = {\mathbb{P}}(v, y) {\mathbb{P}}(b|v)$. Here ${\mathbb{P}}(v, y)$ is the distribution of training data that is unchangeable, while ${\mathbb{P}}(b|v)$ is the distribution that could be optimized.

\subsection{Learning under the IB Framework}

\begin{figure}
	\begin{minipage}[]{0.48\textwidth}
		\includegraphics[width=1.0\textwidth]{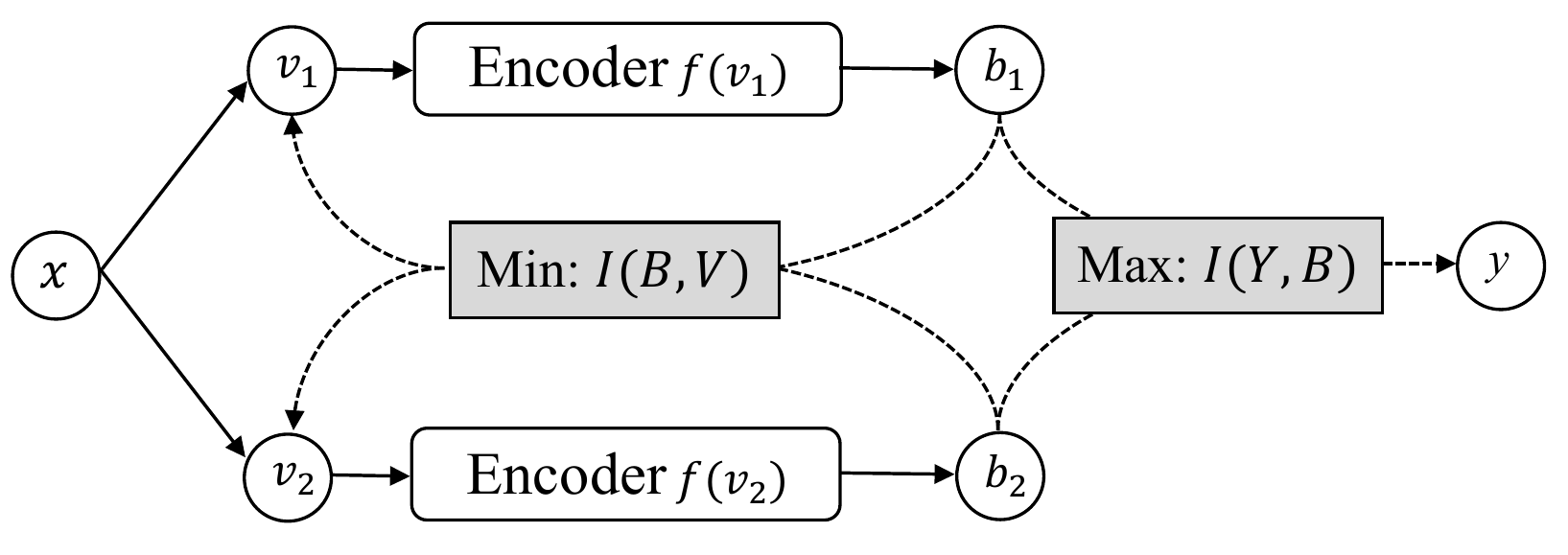} 
	\end{minipage}
	\caption{Illustration of the training procedures of CIBHash. An encoder $f(\cdot)$ is trained via maximizing distinctive semantic information, \textit{i.e.}, $Max:I(Y, B)$; and simultaneously minimizing superfluous information, \textit{i.e.}, $Min:I(B, V)$. } 
	\vspace{-3mm}
	\label{fig:workflow}
\end{figure}

Information bottleneck (IB) is to maximize the mutual information between the representations and output labels, subjecting to some constraint \cite{tishby2000information}. That is, it seeks to maximize the objective function
\begin{equation} \label{IB_objective}
R_{IB} = I(Y, B) - \beta I(B, V),
\end{equation}
where $\beta$ is a Lagrange multiplier that controls the trade-off between the two types of mutual information; and $V$ denotes the random variable of inputs. Obviously, from the perspective of maximizing mutual information, the proposed probabilistic hashing method can be understood under the IB framework by setting the parameter $\beta$ to $0$. It is widely reported that the parameter $\beta$ can control the amount of information that is dropped from the raw inputs $v$, and if an appropriate value is selected, better semantic representations can be obtained \cite{AlemiFD017}. Therefore, we can train the proposed hashing model under the broader IB framework by taking the term $ I(B, V)$ into account.

Instead of directly maximizing $R_{IB}$, we seek to maximize its lower bound due to the computational intractability of $R_{IB}$. For the second term in \eqref{IB_objective}, by definition, we have
$
 I(B, V) = {\mathbb{E}}_{{\mathbb{P}}(v)}\left[KL\left({\mathbb{P}}(b|v) || {\mathbb{P}}(b)\right)\right]. 
$
From the non-negativeness of KL-divergence, it can be easily shown that
\begin{align}
I(B, V) \le {\mathbb{E}}_{{\mathbb{P}}(v)}\left[KL\left({\mathbb{P}}(b|v) || q(b)\right)\right],
\end{align}
where $q(b)$ could be any distribution of $b$. From \eqref{Lce_lower_bound}, we can also get the lower bound for $I(Y, B)$ as  $I(Y, B) \ge -L_{ce} + H(Y)$. Thus, we can obtain a lower bound for $R_{IB}$ as
\begin{equation}
 R_{IB} \ge -L_{ce} -\beta {\mathbb{E}}_{{\mathbb{P}}(v)}\left[KL\left({\mathbb{P}}(b|v) || q(b)\right)\right] + H(Y).
\end{equation}
Since $H(Y)$ is a constant and $L_{ce}$ is exactly the contrastive loss, to optimize the lower bound, we just need to derive the expression for the KL term. By assuming $q(b)$ follows a multivariate Bernoulli distribution $q(b)=Bernoulli(\gamma)$, the expression of KL-divergence $KL\left({\mathbb{P}}(b|v) || q(b)\right)$ can be easily derived to be
\begin{equation}
\resizebox{.89\linewidth}{!}{$
    \begin{aligned}
    KL({\mathbb{P}}(b|v)||q(b)) & = \sum^D_{d=1}  [\sigma(f_\theta(v))]_d  \log \frac{[\sigma(f_\theta(v))]_d}{[\gamma]_d} \\
    & + \sum^D_{i=1}  (1-[\sigma(f_\theta(v))]_d)  \log \frac{1-[\sigma(f_\theta(v))]_d}{1-[\gamma]_d}.
    \label{kl}
    \end{aligned}
    $}
\end{equation}
In our experiments, for a given view $v_1^{(k)}$, the value of $\gamma$ is set by letting $q(b) = p(b|v_2^{(k)})$; and $q(b)$ for the view $v_2^{(k)}$ can be defined similarly. Intuitively, by encouraging the encoding distributions from different views of the same image close to each other, the model can eliminate superfluous information from each view. In this way,  the lower bound of $R_{IB}$ can be maximized by SGD algorithms efficiently. We name the proposed model as CIBHash, standing for \textbf{Hash}ing with \textbf{C}ontrastive \textbf{I}nformation \textbf{B}ottleneck. The architecture is shown in Figure \ref{fig:workflow}.

\section{Experiments}
\subsection{Datasets, Evaluation and Baselines}
\paragraph{Datasets} Three datasets are used to evaluate the performance of the proposed hashing method. \emph{1) CIFAR-10}  is a dataset consisting of $60,000$ images from $10$ classes \cite{Krizhevsky2009Learning}. We randomly select $1,000$ images per class as the query set and $500$ images per class as the training set, and all the remaining images except queries are used as the database. \emph{2) NUS-WIDE}  is a multi-label dataset containing $269,648$ images from $81$ categories \cite{ChuaTHLLZ09}. Following the commonly used setting, the subset with images from the $21$ most frequent categories is used. We select 100 images per class as the query set and use the remaining as the database. Moreover, we uniformly select $500$ images per class from the database for training, as done in \cite{ShenQCY00S020}. \emph{3) MSCOCO} is a large-scale dataset for object detection, segmentation and captioning \cite{lin2014microsoft}. Same as the previous works, a subset of $122,218$ images from $80$ categories is considered. We randomly select $5,000$ images from the subset as the query set and use the remaining images as the database. $10,000$ images from the database are randomly selected for training.


\paragraph{Evaluation Metric} In our experiments, the mean average precision (MAP) at top $N$ is used to measure the quality of obtained hashing codes. Following the settings in \cite{CaoLWY17,ShenQCY00S020}, we adopt MAP@1000 for CIFAR-10, MAP@5000 for NUS-WIDE and MSCOCO. 

\paragraph{Baselines} In this work, we consider the following unsupervised deep hashing methods for comparison: DeepBit \cite{LinLCZ16}, SGH \cite{DaiGKHS17}, BGAN \cite{SongHGXHS18}, BinGAN \cite{ZiebaSET18}, GreedyHash \cite{SuZHT18}, HashGAN \cite{DizajiZSYDH18}, DVB \cite{ShenLS19}, and TBH \cite{ShenQCY00S020}. For the reported performance of baselines, they are quoted from TBH \cite{ShenQCY00S020}. 

\subsection{Training Details}
For images from the three datasets, they are all resized to $224 \times 224 \times 3$. Same as the original contrastive learning \cite{ChenK0H20}, during training the resized images are transformed into different views with transformations like random cropping, random color distortions and Gaussian blur, and then are input into the encoder network. In our experiments, the encoder network $f_\theta(\cdot)$ is constituted by a pre-trained VGG-16 network \cite{SimonyanZ14a} followed by an one-layer ReLU feedforward neural network with 1024 hidden units. During the training, following previous works \cite{SuZHT18,ShenLS19}, we fix the parameters of pre-trained VGG-16 network, while only training the newly added feedfoward neural network. We implement our model on PyTorch and employ the optimizer Adam for optimization, in which the default parameters are used and the learning rate is set to be $0.001$. The temperature $\tau$ is set to $0.3$, and $\beta$ is set to $0.001$.

\begin{table*}[!t]
\caption{MAP comparison with different state-of-the-art unsupervised hashing methods.}
\label{table:map}
\centering
\begin{tabular}{c|c|ccc|ccc|ccc}
\hline
\multirow{2}*{\textbf{Method}}  & \multirow{2}*{\textbf{Reference}}   & \multicolumn{3}{c}{\textbf{CIFAR-10}} & \multicolumn{3}{|c}{\textbf{NUS-WIDE}} & \multicolumn{3}{|c}{\textbf{MSCOCO}}\\

~          & ~         & 16bits & 32bits & 64bits & 16bits & 32bits & 64bits & 16bits & 32bits & 64bits  \\
\hline 
\hline
DeepBit    & CVPR16    & 0.194  & 0.249  & 0.277  & 0.392  & 0.403  & 0.429  & 0.407  & 0.419  & 0.430 \\
SGH        & ICML17    & 0.435  & 0.437  & 0.433  & 0.593  & 0.590  & 0.607  & 0.594  & 0.610  & 0.618 \\
BGAN       & AAAI18    & 0.525  & 0.531  & 0.562  & 0.684  & 0.714  & 0.730  & 0.645  & 0.682  & 0.707 \\ 
BinGAN     & NIPS18    & 0.476  & 0.512  & 0.520  & 0.654  & 0.709  & 0.713  & 0.651  & 0.673  & 0.696 \\
GreedyHash & NIPS18    & 0.448  & 0.473  & 0.501  & 0.633  & 0.691  & 0.731  & 0.582  & 0.668  & 0.710 \\
HashGAN    & CVPR18    & 0.447  & 0.463  & 0.481  & -      & -      & -      & -      & -      & -     \\
DVB        & IJCV19    & 0.403  & 0.422  & 0.446  & 0.604  & 0.632  & 0.665  & 0.570  & 0.629  & 0.623 \\
TBH        & CVPR20    & 0.532  & 0.573  & 0.578  & 0.717  & 0.725  & 0.735  & 0.706  & 0.735  & 0.722 \\
\hline

\textbf{CIBHash}        & \textbf{Ours}      & \textbf{0.590}  & \textbf{0.622}  & \textbf{0.641} & \textbf{0.790}  & \textbf{0.807}  & \textbf{0.815}  & \textbf{0.737}  & \textbf{0.760}  & \textbf{0.775}  \\
\hline
\end{tabular}
\vspace{-2mm}
\end{table*}

\begin{table}[!t]
\caption{MAP comparison with variants of CIBHash.}
\centering
\begin{tabular}{c|c|cccc}
\hline
\multicolumn{2}{c|}{\textbf{Component Analysis}}&  16bits  &  32bits  &  64bits \\
\hline
\hline
\multirow{3}*{CIFAR-10}    &  Naive CL  & 0.493    & 0.574    & 0.606  \\
~                          &  CLHash  &  0.584   &  0.612   &  0.633  \\
~                          &  \textbf{CIBHash}    &  \textbf{0.590}   &  \textbf{0.622}   &  \textbf{0.641}  \\
\hline
\hline

\multirow{3}*{MSCOCO}      &  Naive CL  & 0.666    & 0.712    & 0.737  \\ 
~                          &  CLHash &  0.725   &  0.753   &  0.765  \\
~                          &  \textbf{CIBHash}    &  \textbf{0.737}   &  \textbf{0.760}   &  \textbf{0.775}  \\
\hline
\end{tabular}
\label{table:component}
\vspace{-2mm}
\end{table}

\subsection{Results and Analysis}

\paragraph{Overall Performance} Table \ref{table:map} presents the performances of our proposed model CIBHash on three public datasets with code lengths varying from $16$ to $64$. For comparison, the performance of baseline models is also included. From the table, it can be seen that the proposed CIBHash model outperforms the current SOTA method by a substantial margin on all three datasets considered. Specifically, an averaged improvement of $5.7\%$, $7.8\%$, $3.6\%$ (averaged over different code lengths) on CIFAR-10, NUS-WIDE, and MSCOCO datasets are observed, respectively, when it is compared to the currently best performed method TBH. The gains are even more obvious on the short code case. All of these reveal that the non-reconstruction-based hashing method is really better at extracting important semantic information than the reconstruction-based ones. This also reveals that when contrastive learning is placed under the information bottleneck framework, high-quality hashing codes can be learned. For more comparisons and results, please refer to Supplementary Materials.


\paragraph{Component Analysis}
To understand the influence of different components in CIBHash, we further evaluate the performance of two variants of our model. (\textit{i}) \textbf{Naive CL}: It produces hashing codes by directly binarizing the real-valued representations learned under the original contrastive learning framework using the median value as the threshold. (\textit{ii}) \textbf{CLHash}: CLHash denotes the end-to-end probabilistic hashing model derived from contrastive learning, as proposed in Section \ref{AdaptingCL2Hash}. As seen from Table \ref{table:component}, compared to Naive CL, CLHash improves the averaged performance by $5.2\%$ ,$4.3\%$ on CIFAR-10 and MSCOCO, respectively, which demonstrates the effectiveness of our proposed adaptations on the original contrastive learning, {\textit{i.e.}}, dropping the projection head and enabling end-to-end training. If we compare CIBHash to CLHash, improvements of $0.8\%$ and $1.0\%$ can be further observed on CIFAR-10 and MSCOCO, respectively, which fully corroborates the advantages of considering the CLHash under the broader IB framework.

\begin{table}[!t]
\centering
\caption{MAP comparison with different IB-based methods.}
\label{table:bvae}

\begin{tabular}{l|ccc}
\hline
\multirow{2}*{\textbf{Method}}& \multicolumn{3}{c}{\textbf{CIFAR-10}} \\

~      & 16bits & 32bits & 64bits \\
\hline
$\beta$-VAE    & 0.468  & 0.508  & 0.495  \\
Multi-View $\beta$-VAE &  0.465  & 0.492   &  0.522   \\
\textbf{CIBHash}          &  \textbf{0.590}   & \textbf{0.622}    &  \textbf{0.641}  \\
\hline
\end{tabular}
\vspace{-1mm}
\end{table}

\begin{figure}[!t]
    \centering
	\subfigure{
		\begin{minipage}[b]{0.47\columnwidth}
  	 	\includegraphics[width = 1\columnwidth]{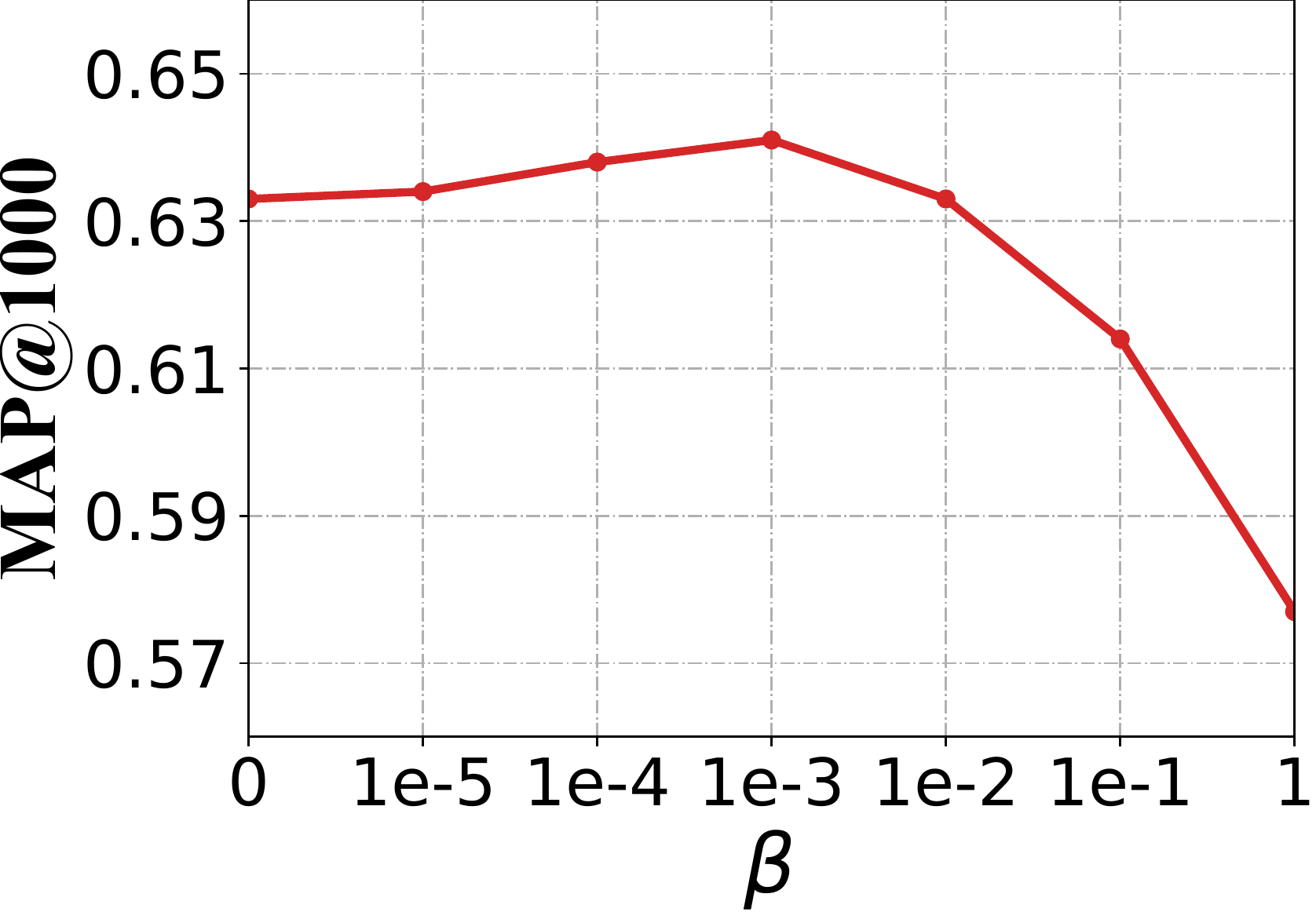}
		\end{minipage}
	}
	\subfigure{
		\begin{minipage}[b]{0.47\columnwidth}
  	 	\includegraphics[width = 1\columnwidth]{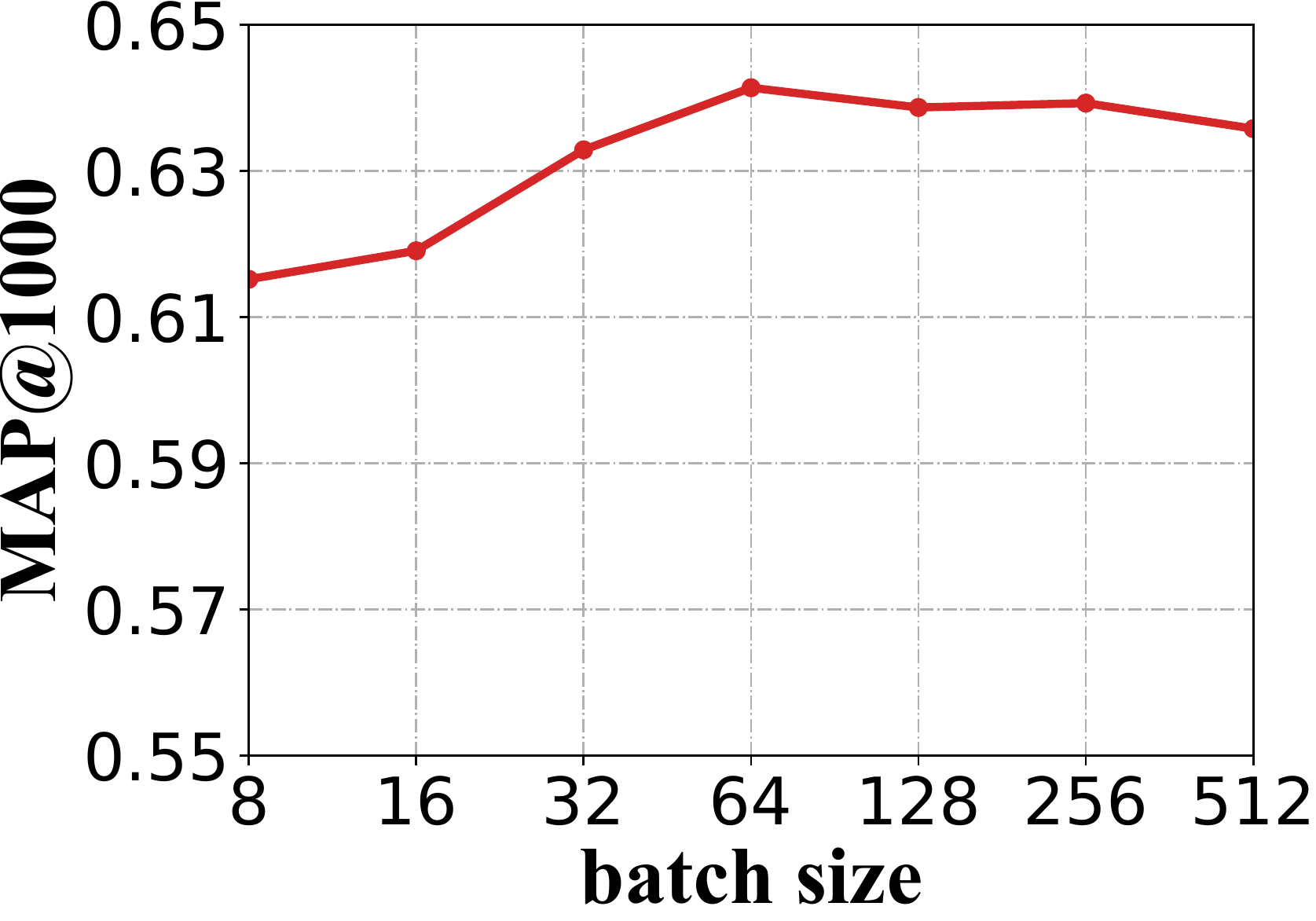}
		\end{minipage}
	}
	\vspace{-1mm}
	\caption{Parameter analysis for the Lagrange multiplier $\beta$ and the batch size with 64-bit hashing codes on CIFAR-10.}
	\label{fig: hypersensitivity}
	\vspace{-0.5mm}
\end{figure}

\paragraph{Parameter Analysis} To see how the key hyperparameters $\beta$ and minibatch size influence the performance, we evaluate the model under different $\beta$ values and minibatch sizes. As shown in the left column of Figure \ref{fig: hypersensitivity}, the parameter $\beta$ plays an important role in obtaining good performance. If it is set too small or too large, the best performance cannot be obtained under either case. Then, due to the observed gains of using large batch sizes in the original contrastive learning, we also study the effect of batch sizes in our proposed CIBHash model. The results are presented in the right column of Figure \ref{fig: hypersensitivity}. We see that as the batch size increases, the performance rises steadily at first and then converges to some certain level when the batch size is larger than $64$.

\paragraph{Comparison with $\beta$-VAE}

$\beta$-VAE can be regarded as an unsupervised representation learning method under the IB framework \cite{AlemiFD017}, where $\beta$ controls the relative importance of reconstruction error and data compression. The main difference between $\beta$-VAE and our model CIBHash is that $\beta$-VAE relies on reconstruction to learn representations, while our model leverages contrastive learning that maximizes the agreement between different views of an image. We evaluate the $\beta$-VAE and multi-view $\beta$-VAE. Table \ref{table:bvae} shows that CIBHash dramatically outperforms both methods. This proves that the non-reconstruction-based method is better at extracting semantic information than the reconstruction-based methods again.



\section{Conclusion}
In this paper, we proposed a novel non-reconstruction-based unsupervised hashing method, namely CIBHash. In CIBHash, we attempted to adapt the contrastive learning to the task of hashing, by which its original structure was modified and a probabilistic Bernoulli representation layer was introduced to the model, thus enabling the end-to-end training.  By viewing the proposed model under the broader IB framework, a more general hashing method is obtained. Extensive experiments have shown that CIBHash significantly outperformed existing unsupervised hashing methods.

\section*{Acknowledgments}
This work is supported by the National Natural Science Foundation of China (No. 61806223, 61906217, U1811264), Key R\&D Program of Guangdong Province (No. 2018B010107005), National Natural Science Foundation of Guangdong Province (No. 2021A1515012299). This work is also supported by Huawei MindSpore.
\newpage
\bibliographystyle{named}
\bibliography{ijcai21}
\clearpage

\begin{appendix}

\section{Comparison with Non-Deep Hashing Methods}
We compare CIBHash with several non-deep unsupervised hashing methods, including ITQ \cite{GongLGP13}, SH \cite{WeissTF08}, SpH \cite{HeoLHCY12}, LSH \cite{Charikar02}, SELVE \cite{ZhuZH14}, AGH \cite{LiuWKC11} and DGH \cite{LiuMKC14}.

Following the setting in \cite{ShenLS19}, we randomly select $100$ images from each class on CIFAR-10 as the query set and use the rest $59,000$ images as the database. $500$ images per class in the database are adopted as the training set. For the reported performance of these baselines, they are quoted from \cite{ShenLS19}. As shown in Table \ref{table:map@all}, our CIBHash consistently outperforms these non-deep hashing methods.

\begin{table}[htbp]
\caption{MAP@All comparison with different non-deep hashing methods on CIFAR-10. }
\label{table:map@all}
\centering
\begin{tabular}{l|cccccc}
\hline
\multirow{2}*{\textbf{Method}}& \multicolumn{3}{c}{\textbf{CIFAR-10}} \\

~      & 16bits & 32bits & 64bits\\
\hline
ITQ    &  0.319  & 0.334  & 0.347 \\
SH     & 0.218  & 0.198  & 0.181 \\
SpH    & 0.229  & 0.253  & 0.283 \\
LSH    & 0.163  & 0.182  & 0.232 \\
SELVE  & 0.309  & 0.281  & 0.239 \\
AGH    & 0.301  & 0.270  & 0.238 \\
DGH    & 0.332  & 0.354    & 0.356 \\
\textbf{CIBHash(Ours)} & \textbf{0.397}  & \textbf{0.421}  & \textbf{0.442} \\
\hline

\end{tabular}

\end{table}

\section{More Performance Comparisons}
\paragraph{Precision@1000} Precision@N is defined as the average ratio of similar instances among the top N returned instances of all queries in terms of Hamming Distance. We also use Precision@1000 to evaluate the performance of our CIBHash on CIFAR-10 and MSCOCO. Table \ref{table:precision} shows that CIBHash consistently outperforms other state-of-the-art methods.

\begin{table}[!t]
\caption{Precision@1000 comparision with different methods. }
\label{table:precision}
\scalebox{0.85}{
\begin{tabular}{l|cccccc}
\hline
\multirow{2}*{\textbf{Method}}& \multicolumn{3}{c}{\textbf{CIFAR-10}} & \multicolumn{3}{c}{\textbf{MSCOCO}} \\

~      & 16bits & 32bits & 64bits & 16bits & 32bits & 64bits \\
\hline

HashGAN     & 0.418  & 0.436  & 0.455  & -      & -      & -     \\
SGH         & 0.387  & 0.380  & 0.367  & 0.604  & 0.615  & 0.637 \\
GreedyHash  & 0.322  & 0.403  & 0.444  & 0.603  & 0.624  & 0.675 \\
TBH         & 0.497  & 0.524  & 0.529  & 0.646  & 0.698  & 0.701 \\ 
\textbf{CIBHash(Ours)}  & \textbf{0.526}  & \textbf{0.570}  & \textbf{0.583}  & \textbf{0.734}  & \textbf{0.767}  & \textbf{0.785} \\
\hline

\end{tabular}
}
\end{table}

\begin{figure*}[htbp]
\centering
	\subfigure[16 bits]{
		\begin{minipage}[b]{0.27\textwidth}
			\includegraphics[width=1.03\textwidth]{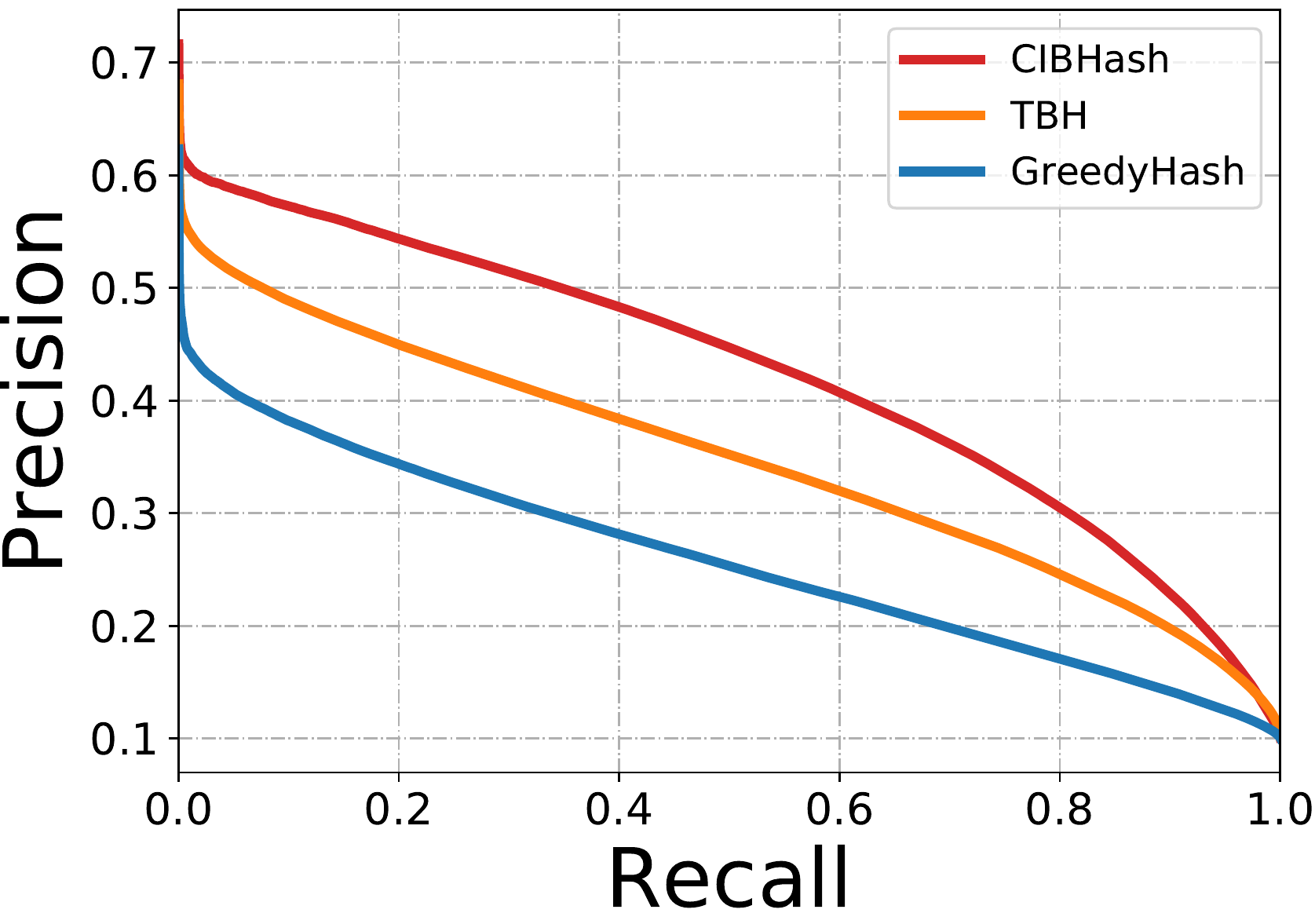}
		\end{minipage}
	}
	\subfigure[32 bits]{
		\begin{minipage}[b]{0.27\textwidth}
   	 	\includegraphics[width=1.03\textwidth]{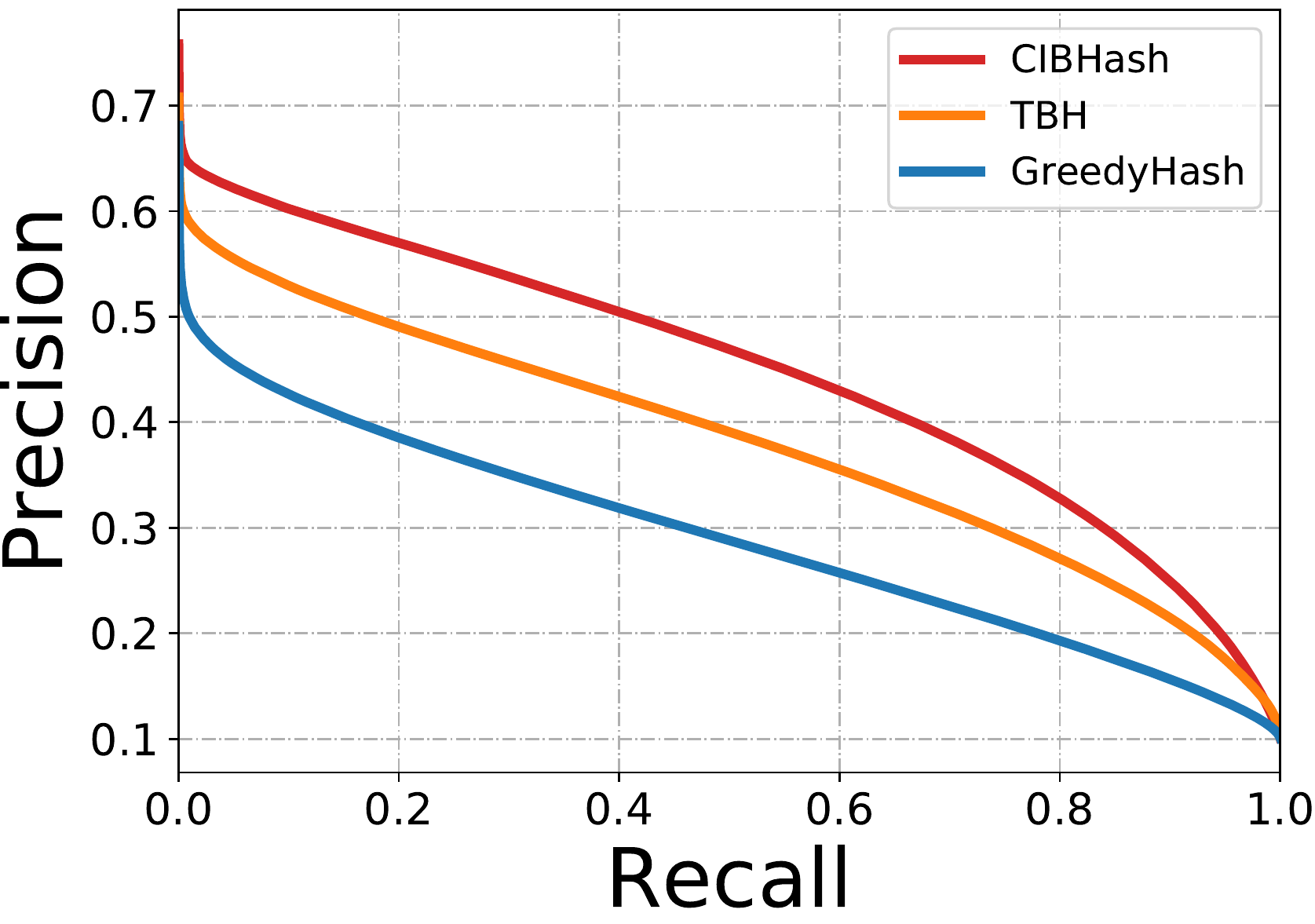}
		\end{minipage}
	}
	\subfigure[64 bits]{
		\begin{minipage}[b]{0.27\textwidth}
   	 	\includegraphics[width=1.03\textwidth]{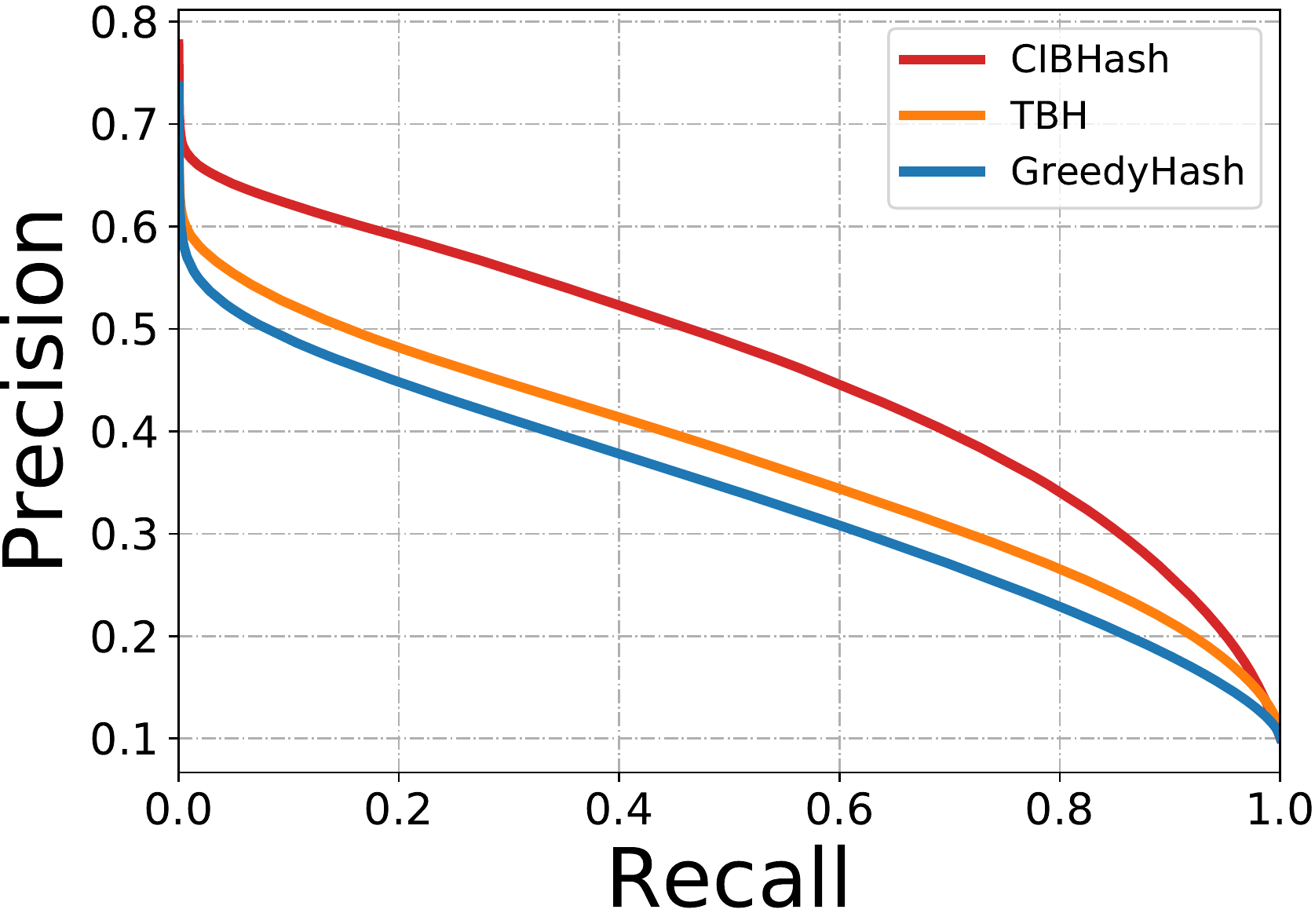}
		\end{minipage}
	}	
	\caption{Precision-Recall curves on CIFAR-10.}
	\label{fig:prcurve}
\end{figure*}

\paragraph{Precision-Recall Curve} The Precision-Recall(P-R) curve is based on hash lookup. Figure \ref{fig:prcurve} shows the P-R curve on CIFAR-10. As can be seen from the result, our retrieval performance is obviously better than that of TBH and GreedyHash.

\section{Influence of Different Temperature}
\begin{figure}[htbp]
\centering
	\begin{minipage}[b]{0.35\textwidth}
		\includegraphics[width=1.03\textwidth]{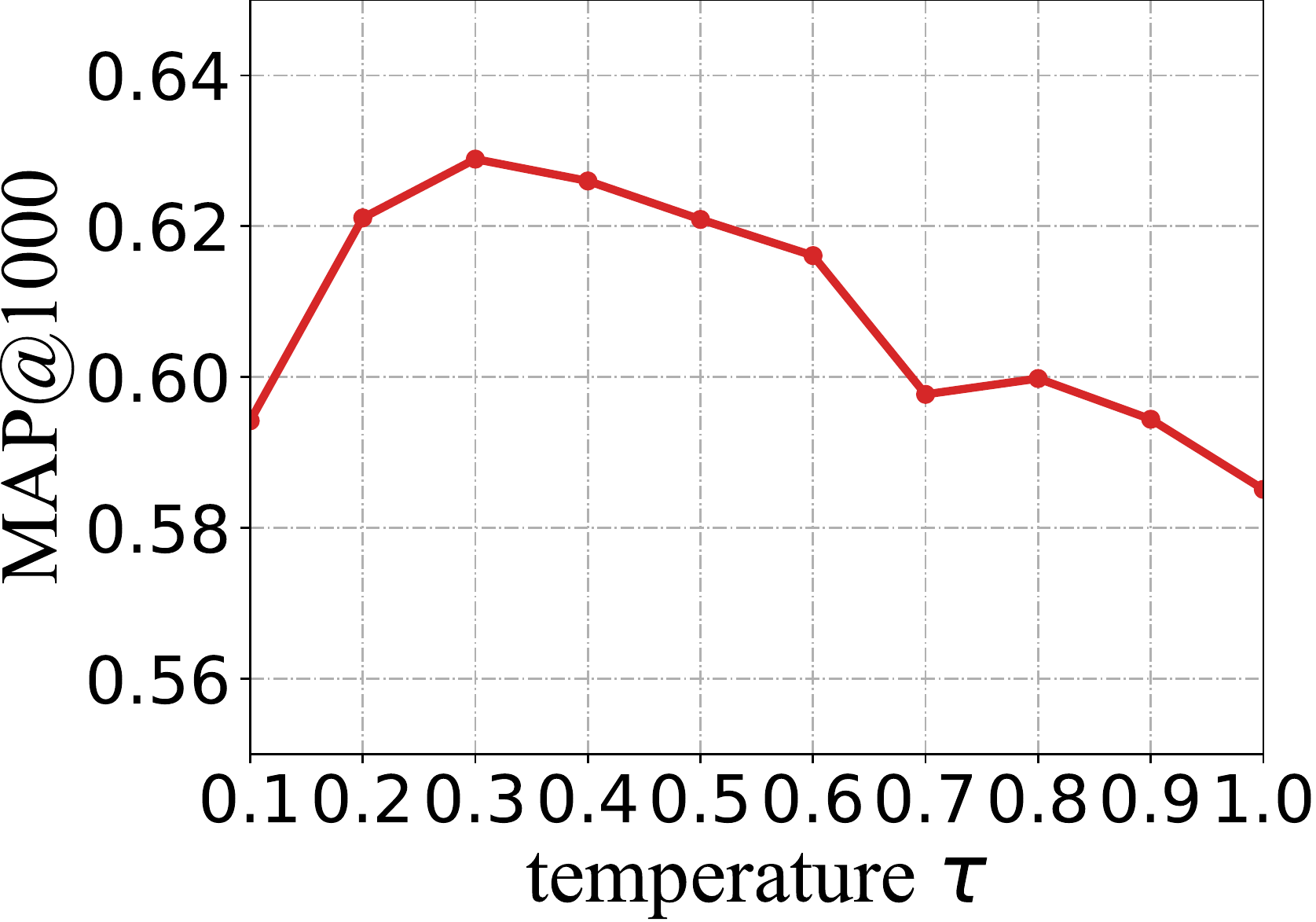} 
	\end{minipage}
	\caption{Model performance with various temperature $\tau$ in terms of 32bits on CIFAR-10.}
    \label{fig:temp}
\vspace{-2mm}
\end{figure}

We further study the influence of the hyperparameter $\tau$. Figure \ref{fig:temp} shows the effect of $\tau$ on CIFAR-10 dataset. We fix $\beta$ to 0.001 and evaluate the MAP by varying $\tau$ between 0.1 and 1. The performance shows that our model performs relatively better with $\tau$ between 0.2 and 0.6. 

\section{Visualization} 
\begin{figure*}[htbp]
\centering
	\begin{minipage}[b]{0.5\textwidth}
		\includegraphics[width=1.03\textwidth]{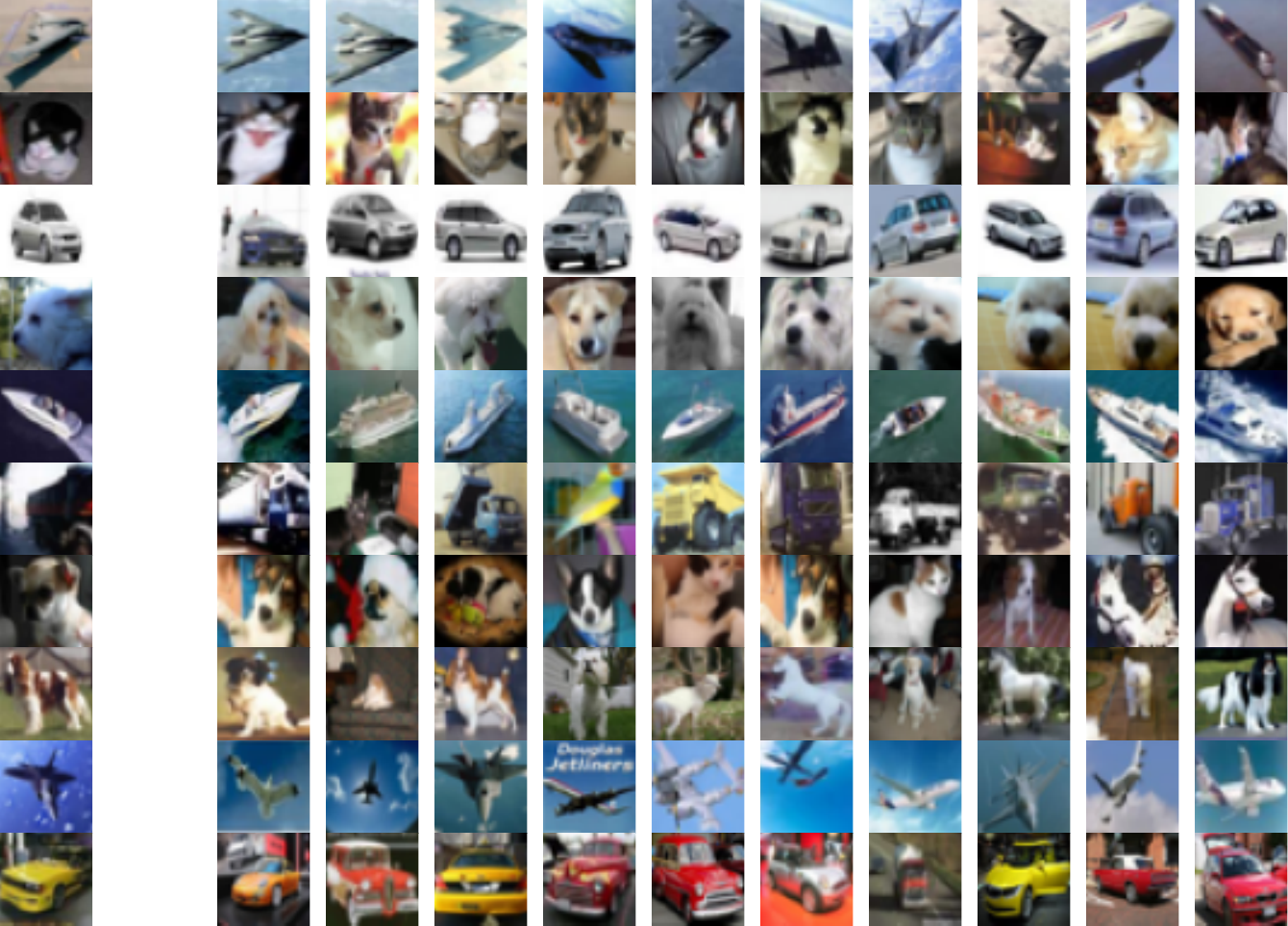} 
	\end{minipage}
	\caption{64-bit retrieval results on CIFAR-10.}
    \label{fig:case}
\end{figure*}

\paragraph{Case Analysis} 
As shown in Figure \ref{fig:case}, most labels of top-10 return images share the same category with queries, demonstrating the superiority of our CIBHash.

\begin{figure*}[htbp]
    \centering
	\subfigure[CIBHash]{
		\begin{minipage}[b]{0.25\textwidth}
			\includegraphics[width = 1\columnwidth]{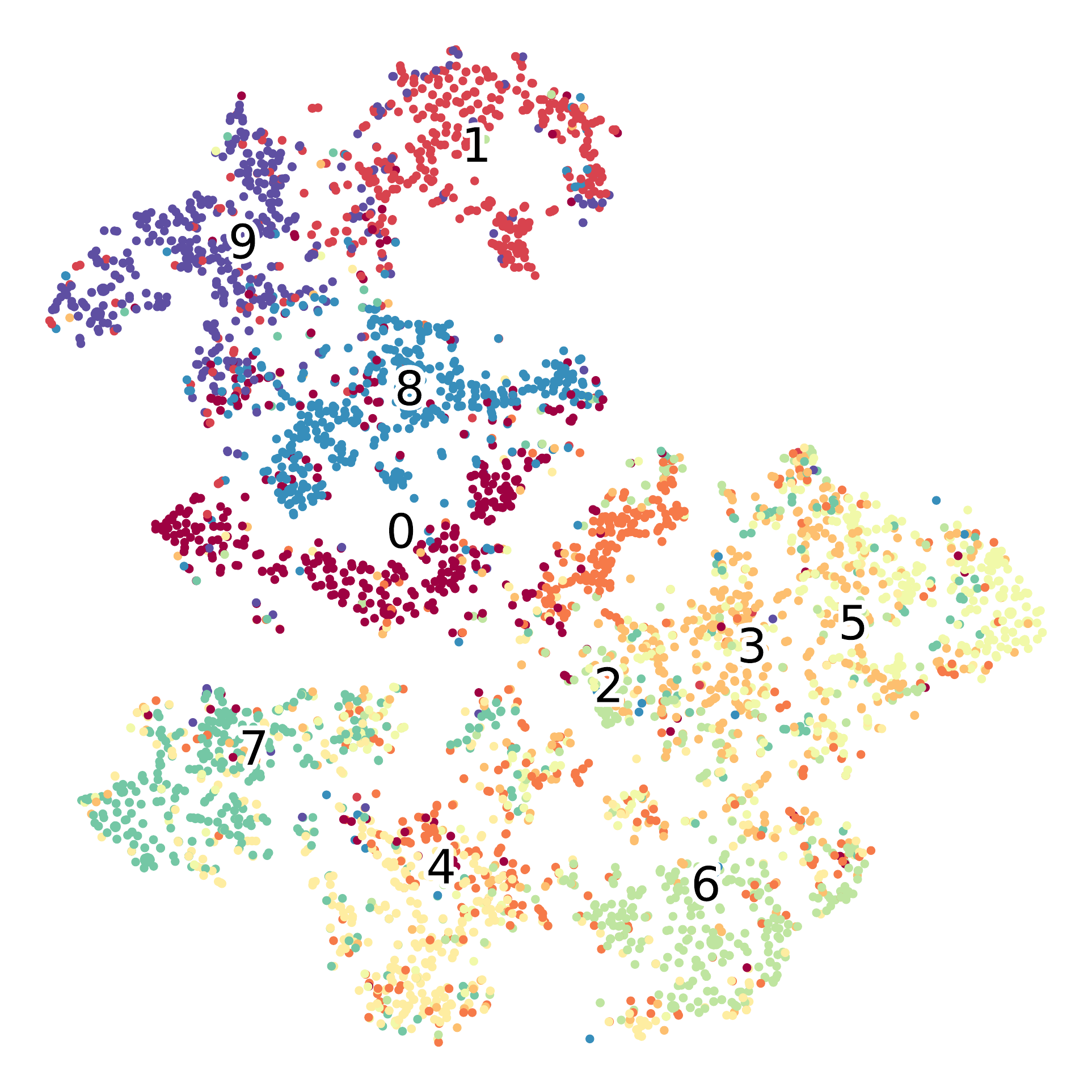}
		\end{minipage}
		\label{fig:ibchvisual}
	}
	\subfigure[TBH]{
		\begin{minipage}[b]{0.3\textwidth}
  	 	\includegraphics[width = 1.0\columnwidth]{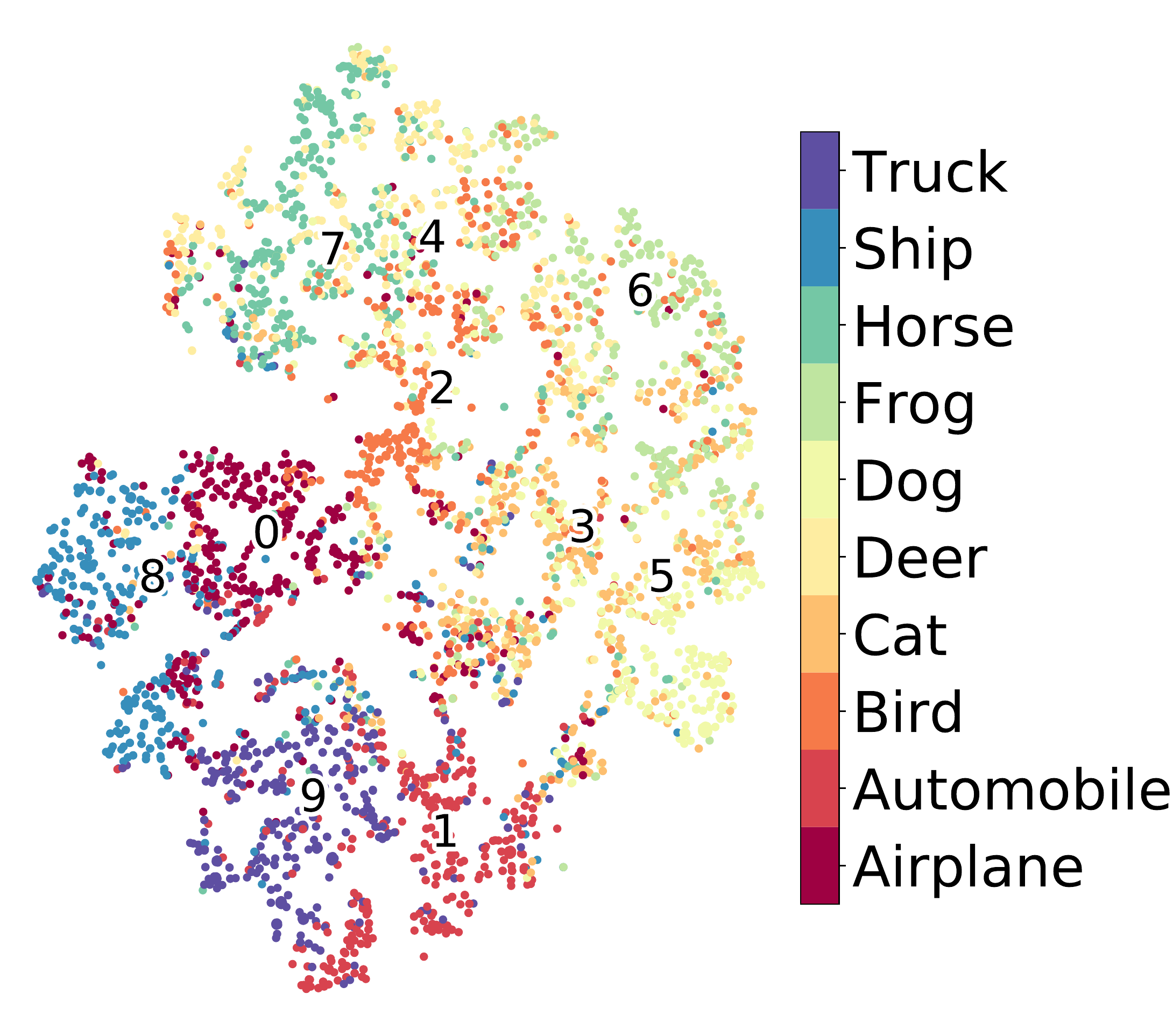}
		\end{minipage}
		\label{fig:tbhvisual}
    }
	\caption{Visualization of the 32-bit hashing codes generated by CIBHash and TBH on CIFAR-10.}
	\label{fig:visualization}
\end{figure*}

\paragraph{t-SNE Projection}
 To intuitively investigate the performance of our CIBHash, we project the 32-bit hashing codes on CIFAR-10 into a 2-dimensional plane. As shown in Figure \ref{fig:visualization}, the hashing codes produced by our CIBHash are more distinguishable when compared with those generated from TBH. 

\vspace{-4mm}

\end{appendix}
\end{document}